%% file: main.tex
\documentclass[a4paper, amsfonts, amssymb, amsmath, reprint, showkeys, nofootinbib, twoside]{revtex4-1}
\usepackage[english]{babel}
\usepackage[utf8]{inputenc}
\makeatletter
\newcommand*{\rom}[1]{\expandafter\@slowromancap\romannumeral #1@}
\makeatother
\usepackage[colorinlistoftodos, color=green!40, prependcaption]{todonotes}
\input{preamble}
\usepackage[pdftex, pdftitle={Article}, pdfauthor={Author}]{hyperref} 
\begin{document}
\title{Geometry Perspective of Estimating Learning Capability of Neural Networks}

\author{Ankan Dutta}
    \email[Correspondence email address: ]{ankan19d@gmail.com}
    \affiliation{ Department of Mechanical Engineering, Jadavpur University, 
  Kolkata 32, India} 
  
\author{Arnab Rakshit}
    \email[Email address: ]{arnabrakshit2008@gmail.com}
    \affiliation{  Artificial Intelligence Lab, Department of Electronics and Telecommunication Engineering, Jadavpur University, 
  Kolkata 32, India}


\begin{abstract}
The paper uses statistical and differential geometric motivation to acquire prior information about the learning capability of an artificial neural network on a given dataset.  The paper considers a broad class of neural networks with generalized architecture performing simple least square regression with stochastic gradient descent (SGD). The system characteristics at two critical epochs in the learning trajectory are analyzed. During some epochs of the training phase, the system reaches \textit{equilibrium} with the generalization capability attaining a maximum. The system can also be \textit{coherent with localized, non-equilibrium states}, which is characterized by the stabilization of the Hessian matrix.  The paper proves that neural networks with higher generalization capability will have a slower convergence rate. The relationship between the generalization capability with the stability of the neural network has also been discussed. By correlating the principles of high-energy physics with the learning theory of neural networks, the paper establishes a variant of the \textit{Complexity-Action conjecture} from an artificial neural network perspective. 
\end{abstract}


\maketitle

\input{sections/section01.tex}  
\input{sections/section02.tex}

\input{sections/section03.tex}
\input{sections/section04.tex}

\input{sections/section05.tex}

\input{sections/section06.tex}
\input{sections/section07.tex}
\input{sections/acknowledgements.tex}

\bibliographystyle{unsrt}  
\bibliography{sections/ref.bib}

\onecolumngrid
\section*{Appendix}

\input{sections/appendix1.tex}

\input{sections/appendix2.tex}

\input{sections/appendix3.tex}

\input{sections/appendix4.tex}

\input{sections/appendix5.tex}

\input{sections/appendix6.tex}

\input{sections/appendix7.tex}

\input{sections/appendix8.tex}

\end{document}

%% file: preamble.tex
\usepackage{amsthm}
\usepackage{mathtools}
\usepackage{physics}
\usepackage{xcolor}
\usepackage{graphicx}
\usepackage[left=23mm,right=13mm,top=35mm,columnsep=15pt]{geometry} 
\usepackage{adjustbox}
\usepackage{placeins}
\usepackage[T1]{fontenc}
\usepackage{lipsum}
\usepackage{csquotes}

%% file: sections/section01.tex
\section{Introduction} \label{sec:introduction}
    Much of the success of Artificial Neural Network (ANN) has been credited to the universal function approximator property of multilayered ANN \cite{HORNIK1991251,10.5555/3086952,kidger2020universal,NIPS2017_7203}. 
ANN with a high number of trainable parameters can approximate any given function with nearly zero training error. Although, this doesn't assure the generalization capacity of ANN on the testing dataset \cite{zhang2016understanding,arora2019finegrained,arora2018stronger}.  In recent work, Zhang \textit{et al} \cite{zhang2016understanding} showed that the neural networks can easily fit random labels with zero training error but the test error or generalization error increases with an increase in randomness in labeling. The generalization error precisely captures the distortion in the underlying meaningful structure of data caused due to random labeling. Hence, generalization plays the most vital role in deciding how well a neural network model learns a dataset. The optimization of the loss function in the training period is an empirically easy task and corresponds to memorization rather than learning. The generalization capability of a model is related to the eigenspectrum of the Hessian matrix of the loss function \cite{chaudhari2016entropysgd}, with wider minima producing better generalization capability \cite{wang2018identifying,sagun2016eigenvalues,chaudhari2016entropysgd}. Laurent \textit{et al} \cite{10.5555/3305381.3305487} on the contrary, argued that sharp minima can have generalization property. Although generalization is not completely understood, it is always assumed that the generalization capability of a neural network is highly dependent on the architecture of the neural network. Using a data-dependent complexity measure, Arora \textit{et al} \cite{arora2019finegrained} showed that the generalization bound is independent of network size. On the other hand intuitively, one can argue that higher training speed will reduce the probability of exploring a better optimal parameter set, thus reducing the generalization capability. The exact dependence of convergence rate and generalization capability is less documented. Here, the paper focuses on the relationship between generalization capability and convergence rate with the architecture of ANN, for a given dataset.

In the past literature, the learning trajectory has been looked up from two different perspectives: a statistical viewpoint and an information-theoretic viewpoint. From a statistical aspect, the learning trajectory represents the evolution of eigenvalues of the hessian matrix of the loss function \cite{jastrzbski2018relation} and evolution of variance of weights implying the generalization capability of a neural network \cite{jastrzbski2018relation}. From an information-theoretic perspective, the learning trajectory (or information flow) consists of the fitting and compression phase of mutual information between layers \cite{shwartzziv2017opening} of a neural network and is dependent on the training dataset and architecture \cite{pmlr-v97-goldfeld19a}. For a particular dataset, different learning trajectory is seen for different neural networks \cite{pmlr-v97-goldfeld19a}.  Here, the paper focuses on the statistical and differential geometric analysis of the learning trajectory. Recent work, Lampinen \textit{et al} \cite{lampinen2018analytic} established the nonlinear dynamics of generalization in deep linear networks. The work by \cite{lampinen2018analytic,saxe2013exact} developed the analytic solutions to the training and testing error of deep networks as a function of training time. However, the work \cite{lampinen2018analytic,saxe2013exact} developed the analytical solutions for a teacher-student network with a fixed architecture where teacher networks generate training data for more complex (or more layered) student network. Here, the paper considers a more generalized architecture framework of ANN with single input-single output data. Learning trajectory manifests the complete dynamical behavior of neural networks. The paper analyses two critical phases of training.  The early phase epoch when the largest (or few large) eigenvalues of the hessian matrix reaches its highest value \cite{jastrzbski2018relation} and the late phase epoch after which the eigenvalues of the hessian matrix completely stabilizes. Focusing only on two critical epochs reduces the complexity of analysis drastically and hence, one can term the analysis of learning trajectory as the analysis of learning capability at the extremal epochs. Learning capability denotes both the effective capability \cite{zhang2016understanding} of a neural network to optimize the loss function (or minimize training error) and also the generalization capability (or minimize testing error) \cite{10.5555/3295222.3295344}.

In 2013, Yan \textit{et al} \cite{YanE4185} argued that the states with local lowest potential or point attractor state represent a particular memory. The stability of these attractor states is crucial for memory storage and retrieval. The paper focuses on the stability and thermodynamics of neural circuits using Lyapunov exponent and probability flux. The analysis exclusively focuses on the stability and thermodynamics of the neural circuit when the ANN reaches its maximum generalization capability. In doing so, the paper establishes that there is always a trade-off between generalization capability and convergence rate, and the system will execute limit cycles or oscillations during maximum generalization capability. The whole analysis is based upon the correspondence between the trajectory of parameters in a Riemannian manifold called Diffusion metric \cite{Fioresi2020} with the trajectory of particles in the space-time where parameters form the generalized coordinate system. The correspondence opens up the application of principles of high-energy physics in the context of learning theory. From this motivation, the paper establishes a variant of the famous Complexity-Action conjecture from an ANN perspective. The correspondence has been made between the complexity of a parameter distribution and the action of parameters on the learning manifold.

The paper constructs the mathematical background in \hyperref[sec: math background]{Section $\text{II}$}, based on which the whole analysis is carried out in \hyperref[sec: metric]{Section $\text{III}$}---\hyperref[sec: metric]{Section $\text{VI}$}. In \hyperref[sec: math background]{Section $\text{II}$} the neural network is represented as a generalized parameterized function, whose loss function is optimized by updating the parameters based on SGD. A matrix called architecture-dataset matrix is introduced which indicates a relation between the architecture of the neural network and the dataset on which it is to be trained. In \hyperref[sec: metric]{Section $\text{II}$} the paper establishes a relationship between Diffusion matrix and the architecture-dataset matrix. The section also proves important learning characteristics of the neural network at two extremal epochs. In \hyperref[sec: stability]{Section $\text{IV}$} the paper analyze the stability using eigenvalues of the architecture-dataset matrix. The relationship between the architecture matrix and the curvature of the diffusion metric is established in \hyperref[sec: ricci]{Section $\text{V}$}. Moreover, in \hyperref[sec: CA]{Section $\text{VI}$} the correlation of the principles of high energy physics to the learning dynamics of neural networks is epitomized by establishing the Complexity-Action correspondence. The major contributions and results of the paper are summarized in \hyperref[sec: conclu]{Section $\text{VII}$}.

%% file: sections/section02.tex
\section{Mathematical  background} \label{sec: math background}
  
  A neural network gets trained on a sampled training dataset and optimizes the loss function iteratively using a learning algorithm. Here stochastic gradient descent (SGD) is employed as the learning algorithm. Let us consider an input data stream $x=\{x_i\}_{i=1}^N$ generated randomly from a Gaussian distribution $P(x)$ to map onto a targeted data stream  $\hat y$. The pair $\{x_i,\hat{y_i}\}_{i=1}^N$ representing $(x_1,\hat{y_1}),(x_2,\hat{y_2}),...,(x_N,\hat{y_N})$ form the training dataset.  The neural network gets trained to learn the mapping of $x\mapsto\hat y$. The neural network produces $y$ as a noisy functional form of $x$ at every training epoch, learning to predict $\hat y$. Mathematically, $y$ can be represented by the equation $1$ as given below

\begin{equation}
y=c(\alpha;x)+\eta
\end{equation}

where $\eta$ denotes a Gaussian noise of standard deviation $\sigma$. The function $c(\alpha;x)$ characterized with parameter $\alpha$ represents the architecture and weight arrangement of the neural network completely. The function $c$ can be represented as the linear combination of its basis function.

\begin{equation}
c(\alpha;x)=\sum_{\mu=1}^K\alpha^\mu \phi_\mu(x) 
\end{equation}

where  $\phi=\{\phi_\mu\}_{\mu=1}^K$ are basis functions of $c$. Here, the co-efficient of basis functions is denoted as the parameters $\alpha=\{\alpha^\mu\}_{\mu=1}^K$. Conventionally, any neural network of layers $L$ is represented by $c=(\phi'_L \circ w_L \circ \phi'_{L-1} \circ ... \circ w_2 \circ \phi'_1 \circ w_1)(x)$ where weights $w=\{w_l\}_{l=1}^L $ and $\alpha$  are related by some function $g$, given by $\alpha^{\mu}=g^\mu (w)$. Activation functions are represented by $\phi'=\{\phi'_l\}_{l=1}^L$. It is important to note that the weights $w$ can be considered as independent of each other but the parameters $\alpha$ are not independent of each other. In fact, the dependency of $\alpha^\mu$ and $\alpha^\nu$ can be given by the Jacobian: 

\begin{equation}
\mathcal{G}^\mu_\nu=\frac{\partial \alpha^\mu}{\partial\alpha^\nu}=
\left\{\begin{tabular}{cc}
$\sum_{l\in\{L(\nu)\}} \frac{\partial g^\mu(w)}{\partial w_l}\frac{\partial w_l}{\partial g^\nu(w)}$  & $\mu\not=\nu$\\
1 & $\mu=\nu$
\end{tabular}
\right\}
\end{equation}

where $\{L(\nu)\}$ is the collection of indexes $l$ for which $\frac{\partial g^\nu(w)}{\partial w^l}\not =0$ which indicates that we are spanning over the weights indexes which will change $g^\nu(w)$. We denote the collection of elements  $\mathcal{G}^\mu_\nu$ as a jacobian matrix $G$, where $G_{\mu\nu}
=\mathcal{G}^\mu_\nu$. The matrix $G$ is not constant to the architecture-dataset pair, $G$ changes with epochs as weights evolves with time.The connectivity between the parameters can be controlled by implicitly changing the element of the matrix $G$. One can change the elements of matrix $G$ inherently by changing the number of layers or changing the dropout rate in the architecture of the neural network. The elements of matrix $G$ signify the dependence of parameters on each other thus it is important in designing of architectures. Note that the matrix $G$ becomes an identity matrix when all the parameters are independent of each other. 

The neural network optimizes the parameter $\alpha$ to the optimal parameter $\bar \alpha$ according to the loss function $f=\frac{1}{N}\sum_{i=1}^N(\hat y_i-y_i)^2$ (here mean squared error) to produce the stream of pair $\{x,\hat y\}$ characterised by $\bar \alpha$. The loss function $f$ can be approximated as an energy term $2\sigma^2E_N$ \cite{bialek2000predictability} with the assumptions of $N \rightarrow \infty$ and neglecting the fluctuations. The energy term $E_N$ can be shown as  

\begin{equation}
\begin{split}
\frac{f}{2\sigma^2}\hspace{2mm}\widetilde{\rightarrow}&E_N(\alpha;\{x,\hat y\})=\frac{1}{2}\\&+\frac{1}{2\sigma^2} \sum_{\mu,\nu=1}^K(\alpha^\mu-\bar \alpha^\mu)A_{\mu\nu}^\infty(\alpha^\nu-\bar \alpha^\nu)
\end{split}
\end{equation}

where $A^\infty$ is a $K \times K$ symmetric matrix representing the architecture and activation functions of the neural network along with the training dataset. So we call $A^\infty$ an architecture-dataset matrix, denoted by

\begin{equation}
A_{\mu\nu}^\infty=\lim_{N\rightarrow \infty}\frac{1}{N}\sum_{i=1}^N\phi_\mu(x_i)\phi_\nu(x_i)=\Big<\phi_\mu(x)\phi_\nu(x)\Big>
\end{equation}

 The variance $\sigma^2$ is constant and an intrinsic property of the dataset sampled from the distribution $P(x)$. Thus the variance $\sigma^2$ is independent of parameters $\alpha$. Considering the basis functions $\phi$ only to be a function of $x$, it is easy to notice that the hessian of loss function w.r.t. parameters $\alpha$  is given by

\begin{equation}
\begin{split}
H_{\zeta\eta}=\frac{\partial^2f}{\partial\alpha^\zeta\partial\alpha^\eta}=&2\sum_{\mu,\nu=1}^K \mathcal{G}^\mu_{\eta\zeta}A^\infty_{\mu\nu}(\alpha^\nu-\bar\alpha^\nu)\\&+2\sum_{\mu,\nu=1}^K\mathcal{G}^\mu_\eta A^\infty_{\mu\nu}\mathcal{G}^\nu_\zeta
\end{split}
\end{equation}

where $\mathcal{G}^\mu_{\eta\zeta}=\frac{\partial^2\alpha^\mu}{\partial\alpha^\eta\partial\alpha^\zeta}$, considering $\alpha$ to be continuous and symmetric in partial derivatives, applying Schwarz's theorem shows $\mathcal{G}^\mu_{\eta\zeta}=\mathcal{G}^\mu_{\zeta\eta}$. One can also note that when the parameters $\alpha^\mu\rightarrow \bar\alpha^\mu$, then the Hessian $H\rightarrow 2GA^\infty G$.

Recent study, \cite{Fioresi2020} defined diffusion matrix and showed that the noise in stochastic gradient descent (SGD) during training is highly anisotropic. The diffusion matrix is basically the co-variance matrix of the gradient of the loss function $f$. Being consistent with the notation of \cite{Fioresi2020}, the diffusion matrix is given by

\begin{equation}
D_{\mu\nu}=\frac{1}{N}\sum_{i=1}^N \frac{\partial f_i}{\partial \alpha^\mu}\frac{\partial f_i}{\partial \alpha^\nu}-\frac{1}{N^2}\sum_{i,j=1}^N \frac{\partial f_i}{\partial \alpha^\mu}\frac{\partial f_j}{\partial \alpha^\nu}
\end{equation}

The diffusion matrix $D$ is a co-variance matrix, thus it is positive semi-definite. Recently \cite{chaudhari2017stochastic} showed that the SGD tend to minimise a separate potential or loss function $\Phi$ (not related to basis functions $\phi$) rather than the original loss function $f$. The potential is defined as

\begin{equation}
\Phi=-\beta^{-1}\log(\rho^{ss}(\alpha))
\end{equation}

where temperature $\beta^{-1} = \frac{\eta}{2b}$, $\eta$ being the learning rate and $b$ is the batch size. $\rho^{ss}(\alpha)$ is the steady state distribution of parameters. The study \cite{Fioresi2020} showed the relation between the potential $\Phi$ and loss function $f$ as $
\nabla \Phi = \widetilde{D} \nabla f 
$ where diffusion metric $\widetilde{D}=I_{K\times K}+\epsilon D$ and $\epsilon$ is a scalar number constraint to $\epsilon<1/\lambda_{D_{max}}$. The diffusion metric is used to derive the geodesic equation which showed the dynamics of weights in anisotropic noise.

%% file: sections/section03.tex
\section{Architecture \& diffusion matrix} \label{sec: metric}
   The relation between architecture-dataset matrix and diffusion matrix shows how the architecture-dataset pair of a neural network governs the stochasticity in SGD. 

\textbf{Identity 1:} Using equation $(7)$ and a similar treatment as \cite{bialek2000predictability} (See \hyperref[A1]{$\text{A}$}), the relation between Architecture matrix $A^\infty$ and Diffusion matrix $D$ is established below 

\begin{equation}
\begin{split}
D_{\mu\nu}\widetilde{\rightarrow}&4[\sum_{p,k,\eta,\zeta}^K \mathcal{G}^\eta_\mu \mathcal{G}^\zeta_\nu (\alpha^p-\bar \alpha^p)\{A_{p\eta \zeta k}^\infty-A_{p \eta}^\infty A_{\zeta k}^\infty\}\\&\quad\quad\quad\quad\quad\quad\times(\alpha^k-\bar \alpha^k)+\sigma^2\sum_{\eta,\zeta}^K\mathcal{G}^\eta_\mu \mathcal{G}^\zeta_\nu A_{\eta\zeta}^\infty] \\
\text{where}\hspace{2mm}& A_{p\eta \zeta k}^\infty=\lim_{N \rightarrow \infty} \frac{1}{N} \sum_{i=1}^N\phi_p(x_i)\phi_\eta(x_i)\phi_\zeta(x_i)\phi_k(x_i)
\end{split}
\end{equation}

The matrix $D$ is evaluated under an approximation for large $N\rightarrow \infty$ and neglecting the fluctuations owing to the symbol $\widetilde{\rightarrow}$ as used in \cite{bialek2000predictability}. Here, an approximated diffusion matrix $D^\infty$ is introduced, which would bring equality to the equation $(9)$ as given below

\begin{equation}
\begin{split}
D^\infty &= 4\sigma^2 GA^\infty G+4C^\infty \\
\text{where}\hspace{2mm} C^\infty_{\mu\nu} &=\sum_{p,k,\eta,\zeta}(\alpha^p-\bar\alpha^p)\mathcal{G}^\mu_\eta\mathcal{G}^\nu_\zeta\{A_{p\eta \zeta k}^\infty-A_{p \eta}^\infty A_{\zeta k}^\infty\}\\&\quad\quad\quad\quad\quad\quad\quad\quad\quad\quad\quad\quad\quad\times(\alpha^k-\bar\alpha^k)
\end{split}
\end{equation}

\textbf{Lemma $1$:} The matrix $C^\infty$ is a symmetric matrix. It is a positive semi-definite matrix if $GA^\infty G$ is positive semi-definite. The elements of matrix $C^\infty$ are given by  

\begin{equation}
\begin{split}
&\sum_{\mu\nu}y^\mu C^\infty_{\mu\nu}y^\nu= 2\sum_{\mu,\nu,\eta,\zeta}V^{ \mu\eta\zeta}V^{ \mu\zeta\eta}\text{var}(M_{\zeta \eta}(x))\\&+\Big[2\sum_{\mu,\nu,\zeta,\eta} y^\nu\mathcal{G}^\nu_\zeta A^\infty_{\eta\zeta}\mathcal{G}^\mu_\eta y^\mu\Big]\Big[2\sum_p (\alpha^p-\bar\alpha^p)^2\text{var}(\phi_p(x))\Big]
\end{split}
\end{equation}

where matrix $V^{\mu p\eta}=y^\mu\mathcal{G}^\mu_\eta(\alpha^p-\bar\alpha^p)$ for any vector $y$ and $M_{\zeta q}(x)=\phi_\zeta(x)\phi_{q}(x)$. (See \hyperref[A2]{$\text{B}$})

Using equation $(10)$ and $(11)$, the approximated diffusion matrix $D^\infty$ can be represented as: 

\begin{equation}
\begin{split}
y^TD^\infty y=&4\sigma^2y^TGA^\infty Gy+8y^TGEGy+ \\&(4\Delta\alpha^TF\Delta\alpha)(4y^TGA^\infty Gy) \geq 0
\end{split}
\end{equation}

where $E$ is a scalar $\Delta\alpha^TY\Delta\alpha$ and $Y_{\mu\nu}=\text{var}(M_{\mu\nu}(x))$ and $\Delta\alpha=\alpha-\bar\alpha$. The elements of matrix $F$ is given by $F_{\mu\nu}=\delta_{\mu\nu}\text{var}(\phi_\mu(x))$. As matrix $Y$ is a variance of the elements of matrix $M$, the scalar $E$ is always positive. Moreover, the matrix $F$ is form of a variance of the elements of the basis function $\phi$. So, the matrix $F$ is positive semi-definite matrix. So, when matrix $GA^\infty G$ is positive semi-definite, the matrix $D^\infty$ is semi-definite, though the converse is not necessarily true. But the inequality comes from the fact the approximated diffusion matrix is also a positive semi-definite matrix similar to the original diffusion matrix. So, the matrix $GA^\infty G$ is constrained to satisfy the inequality which will be analysed later in Theorem $1$. Approximated diffusion matrix $D^\infty$ is always a positive-semi definite matrix as it is the covariance matrix of noise in SGD. But if $GA^\infty G$ is negative semi-definite, then matrix $D^\infty$ can also be indefinite or negative semi-definite. This contradicts the inequality $(12)$, which constrains the learning trajectory. 

We look at the extreme critical epochs when bias $b=\frac{1}{K}\sum_\mu\Delta \alpha_\mu\rightarrow 0$ with maximum parameter variance $\sigma_\alpha^2$ and when bias $b\rightarrow 0$ with minimum parameter variance $\sigma_\alpha^2\rightarrow 0$.  

\textbf{Theorem $1$:} (See \hyperref[A3]{$\text{C}$})
\\
\textit{Part 1}\\
If all of the following conditions are satisfied:\\
\textit{Condition 1:} The matrix $GA^\infty G$ is negative-semi definite i.e. $\max \lambda_{GA^\infty G}<0$ \\
\textit{Condition 2:} Epochs for which bias $b \rightarrow 0$ \\
\textit{Condition 3:} Epochs for which variance of parameters is at its minimum i.e. $\sigma_\alpha^2\rightarrow 0$

Then for these epochs, 
\begin{enumerate}
\item The maximum eigenvalue of the matrix  $GA^\infty G$ i.e. $\max \lambda_{GA^\infty G}\rightarrow 0^-$
\item The matrix $D^\infty$ tends towards null matrix i.e. $D^\infty\rightarrow 0^+$
\item The order of $\Delta\alpha$ is constrained to $\mathcal{O}(\Delta \alpha)\geq \sqrt{\mathcal{O}\Big(\frac{\sigma^2|\max\lambda_{GA^\infty G}|}{\text{det}(F^2)|\max\lambda_{G^2}|-\text{det}(F)|\max\lambda_{GA^\infty G}|}\Big)}$
\item The Hessian of loss function stabilizes i.e. $\lambda_H\rightarrow 0$
\end{enumerate}

\textit{Part 2} \\
Another observation is: \\
Having \textit{Condition 1}, \textit{Condition 2}  satisfied but with \\
\textit{Condition 3:} Epochs for which maximum variance of parameters $\sigma_\alpha^2$ is reached\\
\textit{Condition 4:} We define a matrix $J_{\mu l}=\frac{\partial g^\mu(w)}{\partial w^l}$ and it is full-rank i.e. $\text{rank}(J)=\min\{K,L\}$

Then for these epochs,
\begin{enumerate}
\item The parameters reaches its stationary point $\dot\alpha^\mu\rightarrow 0$ $\forall \mu$ and the system is in equilibrium.
\item The matrix $G,H$ and $D^\infty$ also reaches its stationary point i.e. $\frac{\partial G}{\partial t}\rightarrow 0,\frac{\partial H}{\partial t}\rightarrow 0$ and $\frac{\partial D^\infty}{\partial t}\rightarrow 0$
\item The maximum reached by the variance of parameters is given by $\max\sigma_\alpha^2\sim\frac{1}{\beta\det(A^\infty)}$
\end{enumerate}

In recent literature, \cite{jastrzbski2018relation} showed that the eigenvalue of hessian stabilizes for some epochs, which is an inference of the Theorem $1$: \textit{part 1}. For epochs satisfying Theorem $1$: \textit{Part 1}, there will be isotropic diffusion $D^\infty\rightarrow 0$, so there will be a simple (not stochastic!) gradient descent as Levi-Civita connection coefficients $\Gamma=0$. In other words, the low rank of the diffusion matrix \cite{chaudhari2017stochastic} suggests that the majority of learning is having simple gradient descent for most of the directions, as the parameters are near critical points with minimum variance. Moreover, the matrix $D^\infty\rightarrow 0$ corresponds to the absence of diffusion or coherent process \cite{YanE4185} owing to the non-equilibrium phases and limit cycles shown by \cite{chaudhari2017stochastic}. Yan \textit{et al} \cite{YanE4185} argued that these coherent systems are a crucial part of the stability of the continuous memories in the human brain. Inference drawn from the \textit{part 1} of Theorem $1$ is thus consistent with the works \cite{chaudhari2017stochastic,YanE4185}. The \textit{Part 2} of Theorem $1$ shows that the eigenvalues of hessian at some epochs reaches its maximum value, which is consistent with the work \cite{jastrzbski2018relation}. Our work further shows that the system is in equilibrium at these epochs. 

%% file: sections/section04.tex
\section{Stability analysis of Neural Network} \label{sec: stability}
    We define approximated diffusion metric $\widetilde{D}^\infty=I_{K\times K}+\epsilon D^\infty$, using equation $(13)$. The Lagrangian $\mathcal{L}$ is defined as $\mathcal{L}=\sqrt{\sum_{\mu,\nu}^K\widetilde{D}_{\mu\nu}^\infty\frac{\partial \alpha^\mu}{\partial t} \frac{\partial\alpha^\nu}{\partial t}}- V(\alpha,\dot\alpha)$. The external force for the parameter $\alpha^\mu$ is denoted by $F_\mu=-\frac{\partial V}{\partial \alpha^\mu}$. Recent study \cite{Fioresi2020} used $
\frac{\partial V}{\partial \alpha^\mu}=\frac{\partial }{\partial t}(\frac{\partial f}{\partial \alpha^\mu})
$ to derive the SGD equations for parameters $\alpha$.

\textbf{Identity 2:} The potential $V$ under which the parameters have a geodesics path analogous to SGD is given by (See \hyperref[A4]{$\text{D}$}) 

\begin{equation}
V(\dot\alpha)=\sum_{\mu=1}^K H_{\mu\mu} \int (\dot \alpha^\mu)^2 dt
\end{equation}
where $H$ is the hessian matrix of the function w.r.t. parameters $\alpha$. Now using equation $(10)$ and the definition of Lagrangian, one can relate the action $S=\int L dt$ with the architecture matrix $A^\infty$ given by equation $(14)$

\begin{equation}
S= \int_0^T \Big\{ \sqrt{
\begin{aligned}
&\sum_{\mu=1}^K\dot\alpha^\mu \dot\alpha_\mu+4\epsilon\sum_{\mu,\nu=1}^K \dot\alpha^\mu( C_{\mu\nu}^\infty\\&\quad\quad\quad+ \sigma^2\sum_{\eta,\zeta=1}^K\mathcal{G}^\mu_\eta\mathcal{G}^\nu_\zeta A_{\eta\zeta}^\infty) \dot\alpha^\nu
\end{aligned}
} - V(\dot \alpha) \Big\} dt
\end{equation}

where total training time $T>0$.

Following equation $(14)$, the action is coordinate invariant \cite{MIT2002symmetry} but depends on the path travelled under the learning rule $\dot \alpha$ and is analogous to SGD \cite{Fioresi2020}  given by the geodesic equation 

\begin{equation}
\begin{split}
\frac{\partial\alpha}{\partial t}=&-(I-\epsilon D^\infty)\frac{\partial f}{\partial \alpha}\\
\Rightarrow \frac{\partial\alpha}{\partial t}=&-2GA^\infty(\alpha-\bar\alpha)+2\epsilon D^\infty GA^\infty(\alpha-\bar\alpha)
\end{split}
\end{equation}
The above equations are the dynamical equations governing the parameter evolution with epochs. Considering the matrix $G=I$, the evolution of parameters can be given as 

\begin{equation}
\alpha(t)=\bar\alpha+\exp(-tA^\infty)[\alpha(0)-\bar\alpha]+\mathcal{O}(\epsilon)
\end{equation}

The above equation is equivalent to the evolution of parameters shown in \cite{saxe2013exact} as an exact solution for deep linear networks. Moreover, equation $(15)$ is a generalized form as it can not only capture deep linear networks but any other generalized architecture where the matrix $G$ is a function of $\alpha$. It is evident from equation $(16)$ that the eigenvalues of the matrix $A^\infty$ decides the stability of the neural network. If the matrix $A^\infty$ is positive definite, the system is stable. Whereas, if at least one of the eigenvalue of the matrix $A^\infty$ is negative, a chaotic nature will arise. There can also be limit cycles, which will arise if the matrix $A^\infty$ is singular. Combining the stability analysis from equation $(16)$ for the epochs in \textit{part }$2$ of Theorem $1$ with the matrix $G=I$, it becomes evident that the system reaches unstable equilibrium for negative semi-definite matrix $A^\infty$. Keeping the generalization capability of a neural network i.e. $\max\sigma_\alpha^2$ constant, the quantity $\beta \det(A^\infty)$ remains a constant according to \textit{part} $2$ of Theorem $1$,. Using the dynamical equation $(16)$, the inverse temperature $\beta=\frac{2b}{\eta}$ act as a time-constant factor, or in other words, the inverse of learning rate $1/\eta$ act as a time-constant factor, keeping batch size $b$ constant. Thus, the paper establishes the time-constant factor in equation $(16)$ for a generalized architecture, which corresponds to the time-constant of the dynamical equations used in \cite{saxe2013exact} for teacher-student network.

The problem of searching an optimal architecture-dataset matrix $A^\infty$ (architectures reaching maximum generalization capability with the highest possible convergence rate) takes into account two aspects: generalization capability i.e. $\max \sigma_\alpha^2$ and time-scale of parameter optimization. The generalization capability tends towards infinity when the architecture-dataset matrix tends towards singular $\det A^\infty\rightarrow 0$, using \textit{part 2} of Theorem $1$. On the other hand, the time-scale of parameter optimization is given by the eigenvalues of the architecture-dataset matrix $A^\infty$, which should be a large positive number resulting in a faster optimization rate. Yet, the architecture-dataset matrix must have all positive eigenvalue for the stability of the optimization. To increase the generalization capability $\max \sigma_\alpha^2\rightarrow \infty$, the matrix $A^\infty$ needs to be singular with at least one of the eigenvalue $\lambda\rightarrow 0^-$ to satisfy epochs in \textit{part} $2$ of Theorem $1$. This condition asserts unstable limit cycles in phase space. So, it can be concluded that increasing the generalization capability of a neural network will give rise to limit cycles in the phase portrait of parameters and decreases the convergence or optimization rate.

%% file: sections/section05.tex
\section{Ricci scalar of learning manifold} \label{sec: ricci}

\textbf{Identity $3$:} The Ricci scalar is given by (See \hyperref[A5]{$\text{E}$})

\begin{equation}
R=\frac{\epsilon}{2}\sum_{i,j,p}\Big\{2\frac{\partial^2 D_{ip}^\infty}{\partial\alpha^j\partial \alpha^i}-\frac{\partial^2 D_{ii}^\infty}{\partial\alpha^j\partial \alpha^p}-\frac{\partial^2 D_{pp}^\infty}{\partial\alpha^i\partial \alpha^i}\Big\}
\end{equation}
One can simplify the equation $(17)$ by constraining the matrix $G$.  For architectures having independent parameters $\alpha$, the matrix $G$ becomes identity. Otherwise, in any generalized architecture for which parameters $\alpha$ becomes independent for some epochs, the Ricci scalar is constant given by
\begin{equation}
\begin{split}
R=&2\epsilon\sum_{i,j,p}(2A^\infty_{iipj}+A^\infty_{pi}A^\infty_{ij}+A^\infty_{ip}A^\infty_{pi}-A^\infty_{piij}\\&\quad\quad\quad\quad\quad\quad\quad\quad\quad\quad-A^\infty_{ippi}-2A^\infty_{ii}A^\infty_{pj})
\end{split}
\end{equation}

\textbf{Identity 4:} The Einstein tensor $E$ for architectures with $G=I$ can be related with architecture matrix by (See \hyperref[A6]{$\text{F}$}) 

\begin{equation}
\begin{split}
E_{ik}&=2\epsilon\sum_{j,p}(A^\infty_{jpik}-A^\infty_{jp}A^\infty_{ik}+A^\infty_{ikpj}-A^\infty_{ik}A^\infty_{ip}\\&\quad\quad\quad\quad\quad\quad-A^\infty_{jkip}+A^\infty_{jk}A^\infty_{ip}-A^\infty_{ippk}+A^\infty_{ip}A^\infty_{pk})\\& -\frac{1}{2}\delta_{ik}\sum_{i,j,p}(2A^\infty_{iipj}+A^\infty_{pi}A^\infty_{ij}+A^\infty_{ip}A^\infty_{pi}-A^\infty_{piij}\\&\quad\quad\quad\quad\quad\quad\quad\quad\quad\quad\quad\quad\quad-A^\infty_{ippi}-2A^\infty_{ii}A^\infty_{pj})
\end{split}
\end{equation}

Combining equation $(19)$ with Einstein's field equation $E=K_0T_s$, where $T_s$ is the energy-momentum tensor and $K_0>0$ a constant corresponding to Einstein's constant, it is apparent that the energy-momentum tensor $T_s$ depends on the architecture-dataset matrix. Thus for $G=I$, the mass configuration is constant given by equation $(19)$. The architecture-data pair has a constant corresponding mass configuration (energy density, momentum and stress), which controls the trajectory of parameters in the learning manifold of neural networks. For generalized architectures, as matrix $G$ evolves with time, the corresponding mass configuration dynamically changes in the background manifold. This opens up a possibility to control the learning trajectory of the neural network using the matrix $G$.

%% file: sections/section06.tex
\section{Complexity-Action Conjecture} \label{sec: CA}

The trajectory of parameters in the Riemannian manifold i.e. diffusion metric, can be correlated with the trajectory of a particle in the space-time, and correlations with the principles of high energy physics is a possibility. One such attempt is made in this section by correlating a famous conjecture in the high energy physics called Complexity-Action Conjecture to the learning dynamics of neural network.

Recent literature \cite{Kanwal_2017} defined complexity of a distribution $p$ by the KL-divergence between $p$ and a distribution $p^c$ whose complexity is zero. Complexity zero \cite{Kanwal_2017} is equivalent to the total information flow ($IF$) between past distribution $p^c(t^-)$ and future distribution $p^c(t^+)$ being zero. Total information flow is given by

\begin{equation}
IF(p(t^-)\rightarrow p(t^+)) =\sum_v S(p_v(t^+)|p_v(t^-))-S(p(t^+)|p(t^-))
\end{equation}

where $S$ is the entropy measure and $p_v$ are the distribution of microstates ensemble $v$. Here we defined a class of distribution $p^c$ for which the complexity is $0$. The section applies this defination of complexity on the parameter distribution. 

\textbf{Corollary 2:} Both Dirac-delta distribution $p^{\delta}$ and an equilibrium distribution $p^{ss}$ belongs to the distribution class $p^c$. (See \hyperref[A7]{$\text{G}$})

The time at which the parameter distribution reaches complexity $0$ is taken as $T$.  The  parameter distribution $p(t)$ is characterised with $\alpha(t)$ which learns to reach at critical point $\bar\alpha$ by a learning rule (here SGD).  We define a parameter distribution $p(\alpha)$ whose parameter $\alpha$ is $\epsilon'-$near to the critical points $\bar\alpha$ i.e. $|\alpha'-\bar\alpha|\sim \epsilon'$ at time step $T-\Delta t$, where $\Delta t\rightarrow 0$  and reaching its maximum generalization capacity simultaneously. The complexity of distribution $p$ is equal to the KL divergence between distribution $p$ characterised by $\alpha=\bar\alpha+\epsilon'$ \& $p^{c}$ which is given by

\begin{equation}
C(p,T-\Delta t)=KL (p(\bar\alpha+\epsilon')||p^{c}(\bar\alpha))
\end{equation}

\textbf{Theorem 2:} The action for epochs satisfying \textit{part 1} of Theorem $1$ and \textit{part 2} of Theorem $1$ is given by (See \hyperref[A8]{$\text{H}$})

\begin{equation}
S=2\sigma^2\epsilon C
\end{equation}

A similar argument has been made by relating quantum complexity and action of a certain spacetime region called a Wheeler-DeWitt patch \cite{Brown_2016}. Here the correspondence has been made between complexity of a parameter distribution and action of parameters on the learning manifold. 

%% file: sections/section07.tex
\section{Conclusion} \label{sec: conclu}

A theoretical framework is proposed to estimate the learning capability of a neural network for a particular dataset prior to its training. The paper establishes how the architecture-dataset matrix governs the noise of the SGD during training. At some epochs of the training phase, the system can show absence of noise or diffusion phenomena with localized, non-equilibrium states. Stabilisation of the hessian matrix is associated with such epochs. The system can also reach equilibrium when the variance of parameters reaches its maximum. The paper proves that neural networks will have a slower convergence rate if its generalization capability is large. The stability of the neural network has also been discussed using the eigenvalues of the architecture-dataset matrix. The paper establishes the relationship between the architecture-dataset matrix and the curvature of the diffusion metric. The relationship signifies the correspondence of the trajectory of parameters in the Diffusion metric with the trajectory of particles in the space-time coordinates. Such correspondence correlates the principles of high-energy physics with the learning theory of neural networks, using which the paper establishes the Complexity-Action conjecture from an ANN perspective. Here, the Complexity-Action correspondence has been made between the complexity of the parameter distribution and the action of parameters on the learning manifold. Using the analysis, one can also correlate any neuroscience model (e.g. Hodgkin-Huxley model) with the generalized architecture of the neural network as an extension of the work. The characteristics of learning in the brain thus can be correlated with the learning dynamics of the neural network established in the paper, among which few are discussed using the stability and thermodynamics of the neural network. The correlation can not only help the neuroscience community to understand the mechanism of the human brain, and but also the computer science community to build better neural architectures.

%% file: sections/acknowledgements.tex
\section*{Acknowledgements} \label{sec:acknowledgements}
 A.R. acknowledges the financial support received from Council of Scientific \& Industrial Research (CSIR), Govt. of India.

%% file: sections/appendix1.tex




\subsection{Identity 1 \label{A1}}

\textit{Proof:}\\
Using the definition of diffusion matrix $D$ from equation $(7)$

\begin{multline}
D_{\mu\nu}=\frac{4}{N}\sum_i \{\hat y_i-c(\alpha;x_i)\}\{\hat y_i-c(\alpha;x_i)\} \frac{\partial c(\alpha;x_i)}{\partial \alpha^\mu}\frac{\partial c(\alpha;x_i)}{\partial \alpha^\nu}\\
 -\frac{4}{N^2}\sum_{i,j}\{\hat y_i-c(\alpha;x_i)\}\{\hat y_j-c(\alpha;x_j)\} \frac{\partial c(\alpha;x_i)}{\partial \alpha^\mu}\frac{\partial c(\alpha;x_j)}{\partial \alpha^\nu} 
 \\
 =\frac{4}{N}\sum_i \{\hat y_i-c(\alpha;x_i)\}^2\sum_{\eta,\zeta}\mathcal{G}^\eta_\mu\mathcal{G}^\zeta_\nu \phi_\zeta(x_i)\phi_\eta(x_i)  -\frac{4}{N^2}\sum_{i}\{\hat y_i-c(\alpha;x_i)\}\sum_{\eta}\mathcal{G}^\eta_\mu \phi_\eta(x_i)\sum_{j}\{\hat y_j-c(\alpha;x_j)\}\sum_{\zeta}\mathcal{G}^\zeta_\nu \phi_\zeta(x_i)
\end{multline}

Using a similar approximation argument as \cite{bialek2000predictability}, we let $N \rightarrow \infty$ and neglect the fluctuations.  

\begin{equation}
\begin{split}
\frac{\partial c(\alpha;x_i)}{\partial \alpha^\mu}&=\sum_{\eta}\mathcal{G}^\eta_\mu \phi_\eta(x_i) \hspace{2mm}[\text{as the basis functions are only functions of }x]\\
\lim_{N\rightarrow \infty} \sum_{i=1}^N \hat y_i^2 &=\Big[\sum_{\mu,\nu} \bar\alpha^\mu  \sum_i \phi_\mu(x_i) \phi_\nu(x_i)\bar\alpha^\nu +\sigma^2\Big ] \\
\lim_{N\rightarrow \infty} \sum_{i=1}^N \hat y_i \phi_\mu(x_i) &= \sum_{\nu} \sum_i \phi_\mu(x_i)\phi_\nu(x_i) \bar\alpha^\nu
\end{split}
\end{equation}

Using above relations one can evaluated the diffusion matrix as

\begin{multline}
D_{\mu\nu}^\infty\widetilde{\rightarrow}\frac{4}{N} (\{\sum_{\eta,\zeta,p,q}\mathcal{G}^\zeta_\mu\mathcal{G}^\eta_\nu\bar\alpha^p [\sum_{i}\phi_\zeta(x_i)\phi_\eta(x_i)\phi_p(x_i)\phi_q(x_i)]\bar\alpha^{q}+\sigma^2\sum_{i}\phi_\eta(x_i)\phi_\zeta(x_i)\} \\
-2\sum_{\eta,\zeta,p,q}\mathcal{G}^\zeta_\mu\mathcal{G}^\eta_\nu\alpha^p[\sum_{i}\phi_\eta(x_i)\phi_\zeta(x_i)\phi_p(x_i)\phi_q(x_i)]\bar\alpha^{q}+\sum_{\eta,\zeta,p,q}\mathcal{G}^\zeta_\mu\mathcal{G}^\eta_\nu\alpha^p [\sum_{i}\phi_\eta(x_i)\phi_\zeta(x_i)\phi_p(x_i)\phi_q(x_i)]\alpha^{q} )
\\
-\frac{4}{N^2}\{\sum_{p,\eta}(\alpha^p-\bar \alpha^p)\sum_{i}\mathcal{G}^\eta_\mu \phi_\eta(x_i) \phi_{p}(x_i)\}\{\sum_{q,\zeta}(\alpha^q-\bar \alpha^q)\sum_{j}\mathcal{G}^\zeta_\nu  \phi_{\zeta}(x_j)\phi_{q}(x_j)\}
\\
\Rightarrow \frac{D_{\mu\nu}^\infty}{4}\widetilde{\rightarrow}\sigma^2\frac{1}{N}\sum_{i=1}^N\mathcal{G}^\zeta_\mu\mathcal{G}^\eta_\nu\phi_\zeta(x_i)\phi_\eta(x_i)+\sum_{\eta,\zeta,\eta,p,q}^K\mathcal{G}^\zeta_\mu\mathcal{G}^\eta_\nu(\alpha^p-\bar \alpha^p)\{\frac{1}{N}\sum_{i=1}^N\phi_p(x_i)\phi_\zeta(x_i)\phi_\eta(x_i)\phi_q(x_i)\\
-\frac{1}{N}\sum_{i=1}^N\phi_p(x_i)\phi_\eta(x_i)\frac{1}{N}\sum_{i}^N\phi_\zeta(x_i)\phi_q(x_i)\}(\alpha^q-\bar \alpha^q)
\end{multline}

which can be represented as follows:

\begin{equation*}
\frac{D_{\mu\nu}}{4}\widetilde{\rightarrow}\sum_{p,k,\eta,\zeta}^K \mathcal{G}^\eta_\mu \mathcal{G}^\zeta_\nu (\alpha^p-\bar \alpha^p)\{A_{p\eta \zeta k}^\infty-A_{p \eta}^\infty A_{\zeta k}^\infty\}(\alpha^k-\bar \alpha^k)+\sigma^2\sum_{\eta,\zeta}^K\mathcal{G}^\eta_\mu \mathcal{G}^\zeta_\nu A_{\eta\zeta}^\infty
\end{equation*}

where $4-$ rank tensor $A_{p\eta \zeta q}^\infty=\lim_{N \rightarrow \infty} \frac{1}{N} \sum_{i=1}^N\phi_p(x_i)\phi_\eta(x_i)\phi_\zeta(x_i)\phi_q(x_i)$ and  matrix $A^\infty$ is given by equation $(5)$.

%% file: sections/appendix2.tex
\subsection{Lemma 1\label{A2}}

\textit{Proof:}\\
\begin{align*}
A_{p\mu \nu q}^\infty-A_{p \mu}^\infty A_{\nu q}^\infty &=\lim_{N\rightarrow \infty}\frac{1}{N}\Sigma_{i=1}^N\phi_p(x_i)\phi_\mu(x_i)\phi_\nu(x_i)\phi_q(x_i)-\frac{1}{N} \Sigma_{i=1}^N\phi_p(x_i)\phi_\mu(x_i)\frac{1}{N} \Sigma_{i=1}^N\phi_\nu(x_i)\phi_q(x_i) \\
&=\Big<\phi_p(x)\phi_\mu(x)\phi_\nu(x)\phi_q(x)\Big>-\Big<\phi_p(x)\phi_\mu(x)\Big>\Big<\phi_\nu(x)\phi_q(x)\Big> \\
&=\text{cov}(\phi_\nu(x)\phi_q(x),\phi_p(x)\phi_\mu(x)) \\
&=\text{cov}(\phi_p(x)\phi_\mu(x),\phi_\nu(x)\phi_q(x))^T\\
&= \{A_{q\nu \mu p}^\infty-A_{\nu q}^\infty A_{p \mu}^\infty \}^T
\end{align*}

\begin{align*}
C^\infty_{\mu\nu}&=\Sigma_{p,q}(\alpha^p-\bar\alpha^p)\{A_{p\mu \nu q}^\infty-A_{p \mu}^\infty A_{\nu q}^\infty\}(\alpha^q-\bar\alpha^q)\\
&=\Sigma_{p,q}(\alpha^p-\bar\alpha^p)\{A_{q\nu \mu p}^\infty-A_{p \mu}^\infty A_{\nu q}^\infty\}(\alpha^q-\bar\alpha^q) \text{[using above relation]}\\
&=\Sigma_{p,q}(\alpha^p-\bar\alpha^p)\{A_{p\nu \mu q}^\infty-A_{p \nu}^\infty A_{\mu q}^\infty\}(\alpha^q-\bar\alpha^q) \text{[interchanging dummy variables]}\\
&=C^\infty_{\nu\mu}
\end{align*}

Therefore the matrix $C^\infty$ is a symmetric matrix (though it is obvious from the fact that both the matrix $D^\infty$ and $GA^\infty G$ is symmetric)

For any vector $y$

\begin{equation}
\begin{split}
\sum_{\mu\nu}y^\mu C^\infty_{\mu\nu}y^\nu &=\sum_{\mu,p,q,\nu,\zeta,\eta}y^\mu(\alpha^p-\bar\alpha^p)\mathcal{G}^\zeta_\nu\mathcal{G}^\eta_\mu\{A_{p\eta \zeta q}^\infty-A_{p \eta}^\infty A_{\zeta q}^\infty\}(\alpha^q-\bar\alpha^q)y^\nu\\
&=\sum_{\mu,p,q,\nu,\zeta,\eta}y^\mu(\alpha^p-\bar\alpha^p)\mathcal{G}^\zeta_\nu\mathcal{G}^\eta_\mu\text{cov}(\phi_\zeta(x)\phi_q(x),\phi_p(x)\phi_\eta(x)) (\alpha^q-\bar\alpha^q)y^\nu\\
&=\sum_{\mu,p,q,\nu,\eta,\zeta}V^{\nu q \zeta}\text{cov}(M_{\zeta q}(x),M_{\eta p}(x))V^{\mu p\eta} \\
&=\sum_{\mu,p,q,\nu,\eta,\zeta}\mathcal{G}^\zeta_\nu\mathcal{G}^\eta_\mu(\delta_{\eta \zeta}\delta_{pq}+\delta_{\zeta q}\delta_{p\eta}+\delta_{\zeta\eta}+\delta_{\zeta q}\\&+\delta_{p\eta}+\delta_{p q}) V^{\nu q \zeta}\text{cov}(M_{\zeta q}(x),M_{\eta p}(x))V^{\mu p\eta} \hspace{2mm}[\text{using conditions J}_{1:6}]\\
&=2\sum_{\mu,\nu,\eta,\zeta}V^{ \mu\eta\zeta}V^{ \mu\zeta\eta}\text{var}(M_{\zeta \eta}(x))+4\sum_{\mu,\nu, p,\zeta,\eta} V^{\mu p \eta}\Big<\phi_{\zeta}(x)\Big>\Big<\phi_\eta(x)\Big>\text{var}(\phi_p(x))V^{\nu p \zeta} \hspace{2mm}[\text{using } J_7] \\
&=2\sum_{\mu,\nu,\zeta,\eta}V^{ \mu\eta\zeta}V^{ \mu\zeta\eta}\text{var}(M_{\zeta \eta}(x))\\&+4\sum_{\mu,\nu, p,\zeta,\eta} V^{\mu p \zeta}\Big<\phi_{\zeta}(x)\phi_\eta(x)\Big>\text{var}(\phi_p(x))V^{\nu p\eta} [\text{co variance of basis functions is 0}] \\
&=2\sum_{\mu,\nu,\eta,\zeta}V^{ \mu\eta\zeta}V^{ \mu\zeta\eta}\text{var}(M_{\zeta \eta}(x))+4\sum_{\mu,\nu, p,\zeta,\eta} V^{\mu p \zeta}A^\infty_{\zeta\eta}\text{var}(\phi_p(x))V^{\nu p\eta} \\
&=2\sum_{\mu,\nu,\eta,\zeta}V^{ \mu\eta\zeta}V^{ \mu\zeta\eta}\text{var}(M_{\zeta \eta}(x))+\Big[2\sum_{\mu,\nu,\zeta,\eta} y^\nu\mathcal{G}^\nu_\zeta A^\infty_{\eta\zeta}\mathcal{G}^\mu_\eta y^\mu\Big]\Big[2\sum_p (\alpha^p-\bar\alpha^p)^2\text{var}(\phi_p(x))\Big]
\end{split}
\end{equation}

Here matrix $V^{\mu p\eta}=y^\mu\mathcal{G}^\mu_\eta(\alpha^p-\bar\alpha^p)$ and $M_{\zeta q}(x)=\phi_\zeta(x)\phi_{q}(x)$. 

Conditions $J$ for cross-covariance to be non-zero are
$J_1:(\eta-\zeta)^2+(p-q)^2=0; J_2:(\zeta-q)^2+(\eta-p)^2=0; $ $J_3:\zeta=\eta;$ $J_4:\eta=p;$ $J_5:p=q;$ $J_6:\zeta=q$. Because all the basis functions are independent of each other.

$J_7:$  In literature \cite{10.2307/2286081} showed the co-variance of product of random variables can be given as 

\begin{align*}
\text{cov}(xy, uv) &= \mathrm{E}(x)\mathrm{E}(u)\text{cov}(y, v) + \mathrm{E}(x)\mathrm{E}(v)\text{cov}(y, u) + \mathrm{E}(y)\mathrm{E}(u)\text{cov}(x, v) 
\\&+ \mathrm{E}(y)\mathrm{E}(v)\text{cov}(x, u) + \text{cov}(x, u)\text{cov}(y, v) + \text{cov}(x, v)\text{cov}(y, u)
\end{align*}

where $x,y,u,w$ are random variables. The covariance of product is $0$ for independent $x,y,u,w$. Two variables can be dependent only when they are equal (using condition $J_{1:6}$), thus the covariance is non-zero and given by $\text{cov}(xy,uy)=\mathrm{E}(x)\mathrm{E}(u)\text{var}(y)$

%% file: sections/appendix3.tex
\subsection{Theorem 1\label{A3}}

\textit{Proof:} \\

One can rearrange equation $(12)$ as 

\begin{equation}
4\underset{(a)}{\underbrace{y^TGA^\infty Gy}}\underset{(b)\geq 0}{\Big[\underbrace{\sigma^2+4\Delta\alpha^TF\Delta\alpha}\Big]}+8\Big(\underset{(c)\geq 0}{\underbrace{y^TG\Delta\alpha^TY\Delta\alpha Gy}}\Big)\geq 0
\end{equation}

for all vector $y$. The variable $(a)$ can be positive or negative depending on the definiteness of $GA^\infty G$ matrix. If $(a)\geq 0$, the equation $(26)$ is always true. The equation $(27)$ can be represented for negative semi-definite matrix $y^TGA^\infty Gy<0$ or $\mathrm{R}_y(\lambda_{GA^\infty G})<0$ as:

\begin{equation}
\begin{split}
y^T\frac{2G\Delta\alpha^TV\Delta\alpha G}{\sigma^2+4\Delta\alpha^TF\Delta\alpha}y \geq & \Big|y^TGA^\infty Gy\Big|\\
\Rightarrow \mathrm{R}_y(\frac{2G\Delta\alpha^TY\Delta\alpha G}{\sigma^2+4\Delta\alpha^TF\Delta\alpha})\geq & \Big|\mathrm{R}_y(GA^\infty G)\Big|
\end{split}
\end{equation}

where Rayleigh quotient $\mathrm{R}_y(X)=\frac{y^TXy}{y^Ty}$, as $y$ can be any vector, we drop the subscript $y$ from the Rayleigh quotient. For a negative Rayleigh quotient $\mathrm{R}(X)<0$, the inequality $\min\lambda_X\leq\mathrm{R}(X)\leq \max\lambda_X$ can be expressed as $|\min\lambda_X|\geq|\mathrm{R}(X)|\geq |\max\lambda_X|$. 

For \textit{part 1}, as bias $b\rightarrow 0$ and variance of parameters $\sigma_\alpha^2\rightarrow 0$, equation $(28)$ can be written as 

\begin{equation}
\begin{split}
\lim_{\sigma_\alpha^2\rightarrow 0,b\rightarrow 0}\mathrm{R}(\frac{2G\mathrm{R}_{\alpha_Y}(\lambda_Y) G}{\frac{\sigma^2}{\sigma_\alpha^2+b^2}+4\mathrm{R}_{\alpha_F}(\lambda_F)})\geq & \Big|\mathrm{R}(GA^\infty G)\Big| \hspace{2mm}[\text{Dividing denominator and numerator by }\Delta\alpha^T\Delta\alpha ]\\
\lim_{\epsilon'\rightarrow 0}\epsilon'\geq & \Big|\mathrm{R}(GA^\infty G)\Big| \geq \Big|\max\lambda_{GA^\infty G}\Big|\\
\text{Therefore,}\hspace{2mm}\mathrm{R}(GA^\infty G)\rightarrow 0^-  \hspace{2mm}&\&\hspace{2mm}
\max\lambda_{GA^\infty G}\rightarrow 0^-
\end{split}
\end{equation}

where $\mathrm{R}_{\alpha_Y}(\lambda_Y)=\frac{\Delta\alpha^TY\Delta\alpha}{\Delta\alpha^T\Delta\alpha}$. Now using equation $(10)$, the diffusion matrix $D^\infty\rightarrow 4\sigma^2GA^\infty G$ when $\Delta \alpha\rightarrow 0$. Given the matrix $GA^\infty G$ is negative semi-definite, equation $(12)$ shows that maximum of eigenvalue of diffusion matrix $\mathrm{R}(D^\infty)\leq\max\lambda_{GA^\infty G}\rightarrow 0^-$, a contradiction ! So the order of $\epsilon'$ must be constrained in $\mathcal{O}(\epsilon')>\mathcal{O}(|\max\lambda_{GA^\infty G}|)$ which is important to avoid the contradiction.

Rewriting equation $(27)$, the inequality can be rewritten as 

\begin{equation}
2\mathrm{R}(G^2)\Big[\Delta\alpha^TY\Delta\alpha \Big]\geq |\mathrm{R}(GA^\infty G)|\Big[\sigma^2+4\Delta\alpha^TF\Delta\alpha\Big]
\end{equation}

The order of matrix $Y$ i.e. $\mathcal{O}(Y)=\mathcal{O}(\text{var}(\phi^2))=\mathcal{O}(\text{var}(\phi)^2)=\mathcal{O}(F^2)$. One can represent the inequality in $(28)$ in terms of order magnitude as 

\begin{equation}
\begin{split}
2\frac{\max\lambda_{G^2}}{|\max\lambda_{GA^\infty G}|}\geq\frac{2\mathrm{R}(G^2)}{|\mathrm{R}(GA^\infty G)|}&\geq \frac{\sigma^2+4\Delta\alpha^TF\Delta\alpha}{\Delta\alpha^TY\Delta\alpha}\\
\mathcal{O}\Big(\frac{\max\lambda_{G^2}}{|\max\lambda_{GA^\infty G}|}\Big) &\geq \mathcal{O}\Big(\frac{\sigma^2}{\Delta\alpha^TY\Delta\alpha}\Big)+\mathcal{O}(\frac{1}{\text{det}(F)}) \\
\mathcal{O}\Big(\frac{\max\lambda_{G^2}}{|\max\lambda_{GA^\infty G}|}-\frac{1}{\text{det}(F)}\Big)&\geq  \mathcal{O}\Big(\frac{\sigma^2}{\Delta\alpha^TY\Delta\alpha}\Big) \\
\mathcal{O}(\Delta\alpha^TY\Delta\alpha)&\geq \mathcal{O}\Big(\frac{\sigma^2}{\frac{\max\lambda_{G^2}}{|\max\lambda_{GA^\infty G}|}-\frac{1}{\text{det}(F)}}\Big) \\
\mathcal{O}(\Delta \alpha)&\geq \sqrt{\mathcal{O}\Big(\frac{\sigma^2|\max\lambda_{GA^\infty G}|}{\text{det}(F^2)|\max\lambda_{G^2}|-\text{det}(F)|\max\lambda_{GA^\infty G}|}\Big)}
\end{split}
\end{equation}

Though both $\Delta\alpha$ and $\max\lambda_{GA^\infty G}$ tends towards $0$, the order of $\Delta\alpha$ is still restricted to inequality $(31)$ making the diffusion matrix $D^\infty\rightarrow 0^+$. Moreover using equation $(6)$, the Hessian $H = 2GA^\infty G+2GA^\infty\Delta\alpha$, shows the expected value of $H$ also tends to $\mathrm{R}(H)\rightarrow 0^-$ for $\mathcal{O}(\Delta\alpha)<\mathcal{O}(\det G)$ and the expected value of $\mathrm{R}(H)\rightarrow 0^+$ for $\mathcal{O}(\Delta\alpha)>\mathcal{O}(\det G)$. In both case, the eigenvalue of hessian tends towards $0$.


For \textit{part 2}, consider the phase portrait of parameters with $\alpha$ and $\dot\alpha$ as the state vectors. According to Lioville's theorem, in non-dissipative system the phase flow remains incompressible i.e. the area of phase portait remains constant. In other words, the spread or variance of one state vector competes with the variance of the other state vector owing to the area conservation. Here, as bias $\sigma_\alpha^2$ is maximum, the variance of velocity of parameters $\sigma_{\dot\alpha}^2\rightarrow 0$ according to Lioville's theorem. Moreover, as bias attains its minimum value $b\rightarrow 0$ the expected velocity of parameter $\Big<\dot\alpha\Big>\rightarrow 0$. Combining both mean and variance of the momenta $\dot\alpha$ becoming $0$, almost all of the index $\mu$  of $\dot\alpha^\mu\rightarrow 0$. Using condition $4$ of Theorem $1$, the velocity of parameters can be related as: 

\begin{equation}
\begin{split}
\frac{\partial\alpha^\nu}{\partial t}= & \sum_{l\in\{L(\nu)\}}\frac{\partial g^\nu(w)}{\partial w^l}\frac{\partial w^l}{\partial t}\\
\frac{\partial\alpha}{\partial t}= & J\frac{\partial w}{\partial t}\\
J^\dagger\frac{\partial\alpha}{\partial t}= & \frac{\partial w}{\partial t}\\
\frac{\partial w}{\partial t} = & 0
\end{split}
\end{equation}
where $J^\dagger$ is a left inverse of $J$. Thus the system is in equilibrium. 
From equation $(28)$, considering $\max\sigma_\alpha^2>>\sigma^2$,  

\begin{equation}
\begin{split}
\mathrm{R}(\frac{G\mathrm{R}_{\alpha_Y}(\lambda_Y) G}{2\mathrm{R}_{\alpha_F}(\lambda_F)})\geq & \Big|\mathrm{R}(GA^\infty G)\Big| \\
\mathrm{R}(G\Big\{I\frac{\mathrm{R}_{\alpha_V}(\lambda_V) }{2\mathrm{R}_{\alpha_F}(\lambda_F)}+A^\infty\Big\}G)\geq & 0 \\
\end{split}
\end{equation}

The evolution of matrix $G$ plays an important role in the system. 

\begin{equation}
\begin{split}
\frac{\partial \mathcal{G}^\mu_\nu}{\partial t}=&\sum_{l,p=1}^L\frac{\partial^2 g^\mu(w)}{\partial w^p\partial w^l}\frac{\partial w^l}{\partial g^\nu(w)}\dot w^p+\frac{\partial g^\mu(w)}{\partial w^l}\frac{\partial}{\partial w^p}\Big(\frac{1}{\frac{\partial g^\nu(w)}{\partial w^l}}\Big)\dot w^p
\end{split}
\end{equation}
So for $\dot w=0$, the matrix $G$ has reached a stationary point i.e. $\dot G=0$ as well. Similarly, the diffusion matrix $D^\infty$ becomes stationary and also the hessian matrix is stationary using equation $(10)$ and equation $(6)$, respectively. 

Till now, the parameter $\alpha$ is denoted as a vector whose mean and variance changes with epochs. So, the parameters vector $\alpha$ is collected from one individual trajectory of the parameters across epochs. One can generalize by collecting multiple such trajectories. The parameter vector now generalizes to parameter matrix, each column representing $\alpha^\mu$ $\forall \mu$ as a vector, having a mean of $\Big<\alpha^\mu\Big>$ $\forall\mu$. Each column vector of parameters will have a corresponding expected 'optimal' parameter $\Big<\bar\alpha^\mu\Big>$ $\forall \mu$. Hence, the bias $b=|\Big<\alpha\Big>-\Big<\bar\alpha\Big>|$ is a vector. The variance of parameters $\sigma_\alpha^2$ becomes the co-variance matrix $\Sigma_\alpha$. The matrix $G$ is then defined by $\mathcal{G}^\mu_\nu=\frac{\partial \Big<\alpha^\mu\Big>}{\partial\Big<\alpha^\nu\Big>}$. The distribution of parameters is given by $\mathcal{N}(\Big<\alpha\Big>,\Sigma_\alpha)$. For a single roll-out of parameters $\vec\alpha$ across epochs, the potential $\Phi$ can be written  from equation $(8)$  

\begin{equation}
\begin{split}
\Phi(\vec\alpha)=&-\beta^{-1}\log(\rho^{ss}(\alpha))\\
=&\beta^{-1} \Big[\frac{K}{2}\log |\Sigma_\alpha |+ \frac{K}{2} \log (2 \pi) + \frac{1}{2}  (\vec\alpha-\Big<\alpha\Big>)^T\Sigma^{-1}_\alpha(\vec\alpha-\Big<\alpha\Big>)\Big]\\ 
\Rightarrow \frac{\partial\Phi(\vec\alpha)}{\partial \Big<\alpha^\eta\Big>}=&2\beta^{-1}\sum_{\mu,\nu=1}^K(\vec\alpha-\Big<\alpha^\mu\Big>)\Sigma_\alpha^{-1}(\mu,\nu)\mathcal{G}^\nu_\eta \\
\Rightarrow \nabla \Phi(\vec\alpha)=&2\beta^{-1}G\Sigma_\alpha^{-1}(\vec\alpha-\Big<\alpha\Big>)
\end{split}
\end{equation}

On the other hand, $\nabla\Phi(\alpha)=\widetilde{D^\infty}\nabla f$, where $\nabla f=2GA^\infty(\vec\alpha-\Big<\bar\alpha\Big>)$. As bias $b\rightarrow\vec 0$, the expected value of parameter $\Big<\alpha\Big>\rightarrow\Big<\bar\alpha\Big>$, thus equating two equation 

\begin{equation}
\begin{split}
\beta\widetilde{D^\infty}GA^\infty(\vec\alpha-\Big<\bar\alpha\Big>)=&G\Sigma_\alpha^{-1}(\vec\alpha-\Big<\bar\alpha\Big>)\\
\Rightarrow \beta\Sigma_\alpha G^{-1}\widetilde{D^\infty}GA^\infty=&I \\
\Rightarrow \beta\Sigma_\alpha A^\infty+\epsilon \beta\Sigma_\alpha G^{-1}D^\infty G A^\infty=&I \\
\Rightarrow \det(\Sigma_\alpha ) = \frac{1}{\beta\det(A^\infty) } + \mathcal{O}(\epsilon)
\end{split}
\end{equation}
Reducing the dimensionality, the variance of parameters $\sigma_\alpha^2\sim\det(\Sigma_\alpha)=\frac{1}{\beta\det(A^\infty)}$.

%% file: sections/appendix4.tex
\subsection{Identity 2 \label{A4}}

\textit{Proof:}\\

Evaluating the potential used by \cite{Fioresi2020}, 
\begin{equation}
\begin{split}
\frac{\partial V}{\partial \alpha^\mu}&=\frac{\partial}{\partial t}(\frac{\partial f}{\partial \alpha^\mu})\\
\Rightarrow \frac{\partial V}{\partial t}&=\sum_{\mu} \dot \alpha^\mu \frac{\partial}{\partial t}(\frac{\partial f}{\partial \alpha^\mu}) \hspace{2mm}\text{(following rules of Einstein's summation notation)} \\
\Rightarrow V&=\sum_\mu \int \dot \alpha^\mu \frac{\partial}{\partial t}(\frac{\partial f}{\partial \alpha^\mu}) dt\\ 
 &= \sum_\mu\dot \alpha^\mu \frac{\partial f}{\partial \alpha^\mu}-\sum_\mu \int \ddot \alpha^\mu \frac{\partial f}{\partial \alpha^\mu} dt \\
 &= \sum_\mu\dot \alpha^\mu \frac{\partial f}{\partial \alpha^\mu}-\sum_\mu \int \frac{\ddot \alpha^\mu}{\dot \alpha^\mu} \frac{\partial f}{\partial \alpha^\mu} d\alpha^\mu \\
 &= \sum_\mu\dot \alpha^\mu \frac{\partial f}{\partial \alpha^\mu}-\sum_\mu \Big\{\frac{\partial f}{\partial \alpha^\mu}\dot\alpha^\mu-\frac{\partial^2 f}{\partial\alpha^\mu \partial\alpha^\mu}\int (\dot\alpha^\mu)^2 dt\Big\} \\
 &=\sum_\mu H_{\mu\mu}\int (\dot\alpha^\mu)^2 dt
\end{split}
\end{equation}

where $H$ is the hessian matrix of the loss function. 
If the learning principle of one parameter coordinate $\alpha^\mu$ would depend on $\alpha^\nu$ then the potential $V$ would have been a summation over both dummy variable $\mu,\nu$.

%% file: sections/appendix5.tex
\subsection{Identity 3 \label{A5}}

\textit{Proof:}\\
The Ricci Tensor is given by on the approximated diffusion metric $\widetilde{D}^\infty=I_{K\times K}+\epsilon D^\infty$

\begin{equation}
\begin{split}
R_{ijk}^j&=\frac{\partial \Gamma_{ik}^j}{\partial \alpha^j}-\frac{\partial \Gamma_{jk}^j}{\partial \alpha^i}+\sum_p (\Gamma_{ik}^p \Gamma_{jp}^j- \Gamma_{jk}^p \Gamma_{ip}^j) \\
&=\frac{\epsilon}{2}\sum_p\Big[\frac{\partial}{\partial \alpha^j}\Big\{\delta_{jp}(\frac{\partial D_{pi}^\infty}{\partial \alpha^k}+\frac{\partial D_{pk}^\infty}{\partial \alpha^i}-\frac{\partial D_{ki}^\infty}{\partial \alpha^p})\Big\}-\frac{\partial}{\partial \alpha^i}\Big\{\delta_{jp}(\frac{\partial D_{pj}^\infty}{\partial \alpha^k}+\frac{\partial D_{pk}^\infty}{\partial \alpha^j}-\frac{\partial D_{kj}^\infty}{\partial \alpha^p})\Big\}\Big]+\mathcal{O}(\epsilon^2) \\
&=\frac{\epsilon}{2}\sum_p\Big\{\frac{\partial^2 D_{pi}^\infty}{\partial\alpha^j\partial \alpha^k}+\frac{\partial^2 D_{kp}^\infty}{\partial\alpha^i\partial \alpha^j}-\frac{\partial^2 D_{ki}^\infty}{\partial\alpha^j\partial \alpha^p}-\frac{\partial^2 D_{pp}^\infty}{\partial\alpha^i\partial \alpha^k}\Big\} \\
\Rightarrow R_{ik}&=\frac{\epsilon}{2}\sum_{j,p}\Big\{\frac{\partial^2 D_{pi}^\infty}{\partial\alpha^j\partial \alpha^k}+\frac{\partial^2 D_{kp}^\infty}{\partial\alpha^i\partial \alpha^j}-\frac{\partial^2 D_{ki}^\infty}{\partial\alpha^j\partial \alpha^p}-\frac{\partial^2 D_{pp}^\infty}{\partial\alpha^i\partial \alpha^k}\Big\} \\
\Rightarrow R&=\frac{\epsilon}{2}\sum_{i,k}\delta_{ik}\Big[\sum_{j,p}\Big\{\frac{\partial^2 D_{pi}^\infty}{\partial\alpha^j\partial \alpha^k}+\frac{\partial^2 D_{kp}^\infty}{\partial\alpha^i\partial \alpha^j}-\frac{\partial^2 D_{ki}^\infty}{\partial\alpha^j\partial \alpha^p}-\frac{\partial^2 D_{pp}^\infty}{\partial\alpha^i\partial \alpha^k}\Big\}\Big] +\mathcal{O}(\epsilon^2)\\
&=\frac{\epsilon}{2}\sum_{i,j,p}\Big\{2\frac{\partial^2 D_{ip}^\infty}{\partial\alpha^j\partial \alpha^i}-\frac{\partial^2 D_{ii}^\infty}{\partial\alpha^j\partial \alpha^p}-\frac{\partial^2 D_{pp}^\infty}{\partial\alpha^i\partial \alpha^i}\Big\}
\end{split}
\end{equation}

For matrix $G=I$, the Diffusion matrix is given by 

\begin{equation}
D^\infty(\alpha)=4\sigma^2 A^\infty+4C^\infty(\alpha)
\end{equation}
where $C^\infty_{\mu\nu}=\sum_{p,k} (\alpha^p-\bar \alpha^p)\{A_{p\mu \nu k}^\infty-A_{p \mu}^\infty A_{\nu k}^\infty\}(\alpha^k-\bar \alpha^k)$. So the Ricci Scalar can be represented for these architecture as 

\begin{equation}
\begin{split}
R=&2\epsilon\sum_{i,j,p}\Big\{2\frac{\partial^2 C_{ip}^\infty}{\partial\alpha^j\partial \alpha^i}-\frac{\partial^2 C_{ii}^\infty}{\partial\alpha^j\partial \alpha^p}-\frac{\partial^2 C_{pp}^\infty}{\partial\alpha^i\partial \alpha^i}\Big\} \\
=&2\epsilon\sum_{i,j,p}(2A^\infty_{iipj}+A^\infty_{pi}A^\infty_{ij}+A^\infty_{ip}A^\infty_{pi}-A^\infty_{piij}-A^\infty_{ippi}-2A^\infty_{ii}A^\infty_{pj})
\end{split}
\end{equation}

The Ricci scalar is a constant when matrix $G=I$, in other words the parameters or coordinates are independent of each other.

%% file: sections/appendix6.tex
\subsection{Identity 4 \label{A6}}

\textit{Proof:}\\
The Einstein tensor can be given for architectures with $G=I$ as
\begin{equation}
\begin{split}
G_{ik}&=\frac{\epsilon}{2}\sum_{j,p}\Big\{\frac{\partial^2 D_{pi}^\infty}{\partial\alpha^j\partial \alpha^k}+\frac{\partial^2 D_{kp}^\infty}{\partial\alpha^i\partial \alpha^j}-\frac{\partial^2 D_{ki}^\infty}{\partial\alpha^j\partial \alpha^p}-\frac{\partial^2 D_{pp}^\infty}{\partial\alpha^i\partial \alpha^k}\Big\}-\frac{1}{2}\delta_{ik}\sum_{i,j,p}\Big\{2\frac{\partial^2 D_{ip}^\infty}{\partial\alpha^j\partial \alpha^i}-\frac{\partial^2 D_{ii}^\infty}{\partial\alpha^j\partial \alpha^p}-\frac{\partial^2 D_{pp}^\infty}{\partial\alpha^i\partial \alpha^i}\Big\} +\mathcal{O}(\epsilon^2) \\
&=2\epsilon\sum_{j,p}(A^\infty_{jpik}-A^\infty_{jp}A^\infty_{ik}+A^\infty_{ikpj}-A^\infty_{ik}A^\infty_{ip}-A^\infty_{jkip}+A^\infty_{jk}A^\infty_{ip}-A^\infty_{ippk}+A^\infty_{ip}A^\infty_{pk})\\& -\frac{1}{2}\delta_{ik}\sum_{i,j,p}(2A^\infty_{iipj}+A^\infty_{pi}A^\infty_{ij}+A^\infty_{ip}A^\infty_{pi}-A^\infty_{piij}-A^\infty_{ippi}-2A^\infty_{ii}A^\infty_{pj})
\end{split}
\end{equation}

%% file: sections/appendix7.tex
\subsection{Corollary 2 \label{A7}}

For dirac delta function $p^\delta$, there is only one microstate $\nu=1$. So $IF=C=0$. 
On the other hand, the future equilibrium distribution $p^{ss}(t^+)$ can be completely predicted by the past distribution $p^{ss}(t^-)$ thus $H(p^{ss}(t^+)|p^{ss}(t^-))=0$. Moreover as the distribution is in equilibrium, probability of micro states $v$ are stationary $H(p_v^{ss}(t^+)|p_v^{ss}(t^-))=0$ $\forall v$. Thus the $IF$ and complexity of any distribution in equilibrium is zero.

%% file: sections/appendix8.tex
\subsection{Theorem 2 \label{A8}}

\textit{Proof:}\\
For $\Delta t\rightarrow 0$,
\begin{equation}
\begin{split}
C(T-\Delta t)&=D_{KL} (p(\bar\alpha+\epsilon')||p^{c}(\bar\alpha)) \\
&=\frac{1}{2}\sum_{\mu,\nu}^K \Delta\alpha^\mu\Big[\frac{1}{N}\sum_{i=1}^N\nabla_\mu \log p^t((x_i,y_i) | \alpha) \nabla_\nu \log p^t((x_i,y_i) |\alpha)\Big]_{\mu\nu}\Delta\alpha^\nu\\
&=\frac{1}{4\sigma^2} \sum_{\mu,\nu}^K\Delta\alpha^\mu\Big[\frac{1}{N}\sum_{i=1}^N\nabla_\mu f_i \nabla_\nu f_i\Big]_{\mu\nu} \Delta\alpha^\nu\\
&=\frac{1}{\sigma^2}\sum_{p,k,\eta,\zeta,\mu,\nu}^K\mathcal{G}^\eta_\mu\mathcal{G}^\zeta_\nu \Delta\alpha^\mu(\alpha^p-\bar \alpha^p) A_{p\eta\zeta k}^\infty (\alpha^k-\bar \alpha^k)\Delta\alpha^\nu+\sum_{\zeta,\eta,\mu,\nu}^K\Delta\alpha^\mu \mathcal{G}^\eta_{\mu}\mathcal{G}^\zeta_\nu A_{\eta\zeta}^\infty \Delta\alpha^\nu  \hspace{2mm}\Big[\text{see equation $(22-24)$}\Big] \\
\Rightarrow \frac{\partial C}{\partial t}\Delta t&=\frac{1}{\sigma^2}\sum_{p,k,\eta,\zeta,\mu,\nu}^K\mathcal{G}^\eta_\mu\mathcal{G}^\zeta_\nu \Delta\alpha^\mu(\alpha^p-\bar \alpha^p) A_{p\eta\zeta k}^\infty (\alpha^k-\bar \alpha^k)\Delta\alpha^\nu+\sum_{\zeta,\eta,\nu,\mu}^K\Delta\alpha^\mu \mathcal{G}^\eta_{\mu}\mathcal{G}^\zeta_\nu A_{\eta\zeta}^\infty \Delta\alpha^\nu\hspace{2mm}[C(T)=0]
\end{split}
\end{equation}

Rewriting an approximation of equation $(14)$ using the definition of Lagrangian $L$ for small $\epsilon<<1$ and $\Delta t\rightarrow 0$.  

\begin{equation}
\begin{split}
Ldt&= \sum_{\mu}^Kd\alpha^\mu d\alpha_\mu+\sum_{\mu,\nu}^K\frac{\epsilon}{2}D_{\mu\nu}^\infty d\alpha_\mu d\alpha_\nu-V(\dot\alpha)dt\\
&= \sum_{\mu}^Kd\alpha^\mu d\alpha_\mu+2\sigma^2\epsilon\frac{\partial C}{\partial t}dt-2\epsilon\sum_{\mu,\nu,p,k,\eta,\zeta}^K\mathcal{G}^\eta_\mu\mathcal{G}^\zeta_\nu d\alpha^\mu(\alpha^p-\bar \alpha^p) A_{p\eta}^\infty A_{\zeta k}^\infty (\alpha^k-\bar \alpha^k)d\alpha^\nu- V(\dot \alpha) dt \\
\frac{dS}{dt}&=\sum_{\mu}^K \dot\alpha^\mu \dot\alpha_\mu dt+2\sigma^2\epsilon\frac{\partial C}{\partial t}-2\epsilon\sum_{\mu,\nu,p,k,\eta,\zeta}^K\mathcal{G}^\eta_\mu\mathcal{G}^\zeta_\nu \dot\alpha^\mu(\alpha^p-\bar \alpha^p) A_{p\eta}^\infty A_{\zeta k}^\infty (\alpha^k-\bar \alpha^k)\dot\alpha^\nu- V(\dot \alpha)  
\end{split}
\end{equation}

For $\alpha \rightarrow \bar\alpha$ and $dt\rightarrow 0$, the rate of change of action can be equated with 

\begin{equation}
\begin{split}
\frac{dS}{dt}=2\sigma^2\epsilon\frac{\partial C}{\partial t}-\sum_\mu^KH_{\mu\mu}\int( \dot\alpha^\mu)^2dt \\
\end{split}
\end{equation}

Epochs satisfying \textit{part 1} of Theorem $1$ , the hessian matrix tends toward a null matrix, thus $H_{\mu\mu}=0$ $\forall \mu$  which makes the equation $(44)$ satisfy $S=2\sigma^2\epsilon C$. 
Epochs satisfying \textit{part 2} of Theorem $1$, the system reaches equilibrium $\dot\alpha\rightarrow 0$ thus  making the equation $(44)$ satisfy $S=2\sigma^2\epsilon C$.